\documentclass[a4paper,10pt,twocolumn]{article}

\usepackage[margin=2.5cm]{geometry}

\usepackage[bookmarks=false]{hyperref}
\usepackage{amsmath,amssymb,amsfonts}
\usepackage{array}
\usepackage[caption=false,font=normalsize,labelfont=sf,textfont=sf]{subfig}
\usepackage{graphicx}
\usepackage{bm}
\usepackage{multirow}
\usepackage{xcolor}

\newcommand{\ve}[1]{\mbox{\boldmath $ #1 $}}

\def\BibTeX{{\rm B\kern-.05em{\sc i\kern-.025em b}\kern-.08em
    T\kern-.1667em\lower.7ex\hbox{E}\kern-.125emX}}

\title{Force Generative Imitation Learning: Bridging Position Trajectory and Force Commands through Control Technique\thanks{
This work was supported by JSPS KAKENHI Grant Number 24K00905,
JST, PRESTO Grant Number JPMJPR24T3, and JST ALCA-Next, Grant Number JPMJAN24F1. This study was based on the results
obtained from the JPNP20004 project subsidized by the New Energy and
Industrial Technology Development Organization (NEDO).
}}

\author{
    Hiroshi Sato\textsuperscript{1},
    Toshiaki Tsuji\textsuperscript{2},
    Sho Sakaino\textsuperscript{3}\thanks{Corresponding author: sakaino@iit.tsukuba.ac.jp} \\[0.5em]
    \small \textit{\textsuperscript{1}Graduate School of Systems and Information Engineering, University of Tsukuba, Japan} \\
    \small \textit{\textsuperscript{2}Graduate School of Science and Engineering, Saitama University, Japan} \\
    \small \textit{\textsuperscript{3}Department of Intelligent Interaction Technologies, Institute of Systems and Information Engineering, University of Tsukuba, Japan}
}

\date{}

\begin{document}

\maketitle

\begin{abstract}
In contact-rich tasks, while position trajectories are often easy to obtain, appropriate force commands are typically unknown.
Although it is conceivable to generate force commands using a pretrained foundation model such as Vision-Language-Action (VLA) models, force control is highly dependent on the specific hardware of the robot, which makes the application of such models challenging.
To bridge this gap, we propose a force generative model that estimates force commands from given position trajectories.
However, when dealing with unseen position trajectories, the model struggles to generate accurate force commands.
To address this, we introduce a feedback control mechanism.
Our experiments reveal that feedback control does not converge when the force generative model has memory.
We therefore adopt a model without memory, enabling stable feedback control.
This approach allows the system to generate force commands effectively, even for unseen position trajectories, improving generalization for real-world robot writing tasks.
\end{abstract}

\vspace{1em}
\noindent\textbf{Keywords:} Imitation Learning, Learning from Demonstration, Motion Generation, Force Control, Machine Learning

\begin{figure}[h]
\centering
    \includegraphics[width=0.9\linewidth]{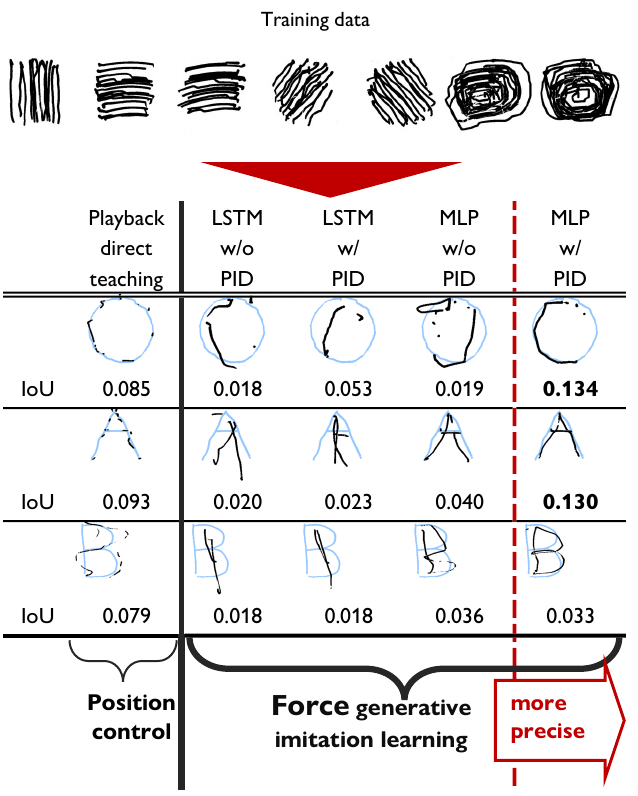}
    \caption{Accuracy of character writing improved by the proposed method}
    \label{fig:Character_writing_tasks_improved_in_accuracy_by_the_proposed_method}
\end{figure}

\section{Introduction}
In recent years, due to labor shortages, robots are expected to perform tasks traditionally carried out by humans.
However, these robots primarily perform predefined actions and need to be reprogrammed when the environment changes
\cite{survey_on_humanrobot}\cite{Industrial_Robot_Control_Trends}.
Therefore, it is difficult for robots to completely replace human workers unless they can adapt to changes in the environment.

To enable robots to adapt their actions to the environment, recent studies have explored the use of machine learning \cite{weinberg2024survey}.
One promising approach is imitation learning, which has the ability to perform complex tasks \cite{pmlr-v205-zhu23a}\cite{zhao2023learningfinegrainedbimanualmanipulation}\cite{wang2023mimicplay}\cite{fu2024mobile}\cite{octo_2023}.
Imitation learning is a type of supervised learning in which a machine learning model learns from experts' demonstrations, resulting in high sampling efficiency.
However, it has a limitation in contact-rich tasks, where it can only handle very static motions.

To address this issue, imitation learning using force control is attracting attention.
In particular, it has been shown that teaching motions using bilateral control, which is a controller that transmits reaction forces from remote environments, enables the acquisition of human-level manipulation skills in the environment \cite{saigusa_ieeeaccess_2022}\cite{akagawa_jia_2023}\cite{yamane2023soft}\cite{buamanee2024bi}\cite{Masato_Kobayashi202524004380}.
The effectiveness of bilateral control has also been demonstrated in practice, as evidenced by its use by the winning team in the ICRA 2024 Food Topping Challenge \cite{Inami05062025}.
Bilateral control makes it possible to collect data on the force perception of a human operator, which is difficult to do with direct teaching
\cite{directteachingkushida2001human}\cite{ablettMultimodalForceMatchedImitation2023}.

In motion collection via direct teaching \cite{fu2024incontextimitationlearningnexttoken}, only the response (state) of a robot can be obtained.
In contrast, bilateral control, which is a teleoperation framework consisting of leader and follower robots, enables the independent collection of both command (action) and state because the action of the robot corresponds to the leader's response while the state corresponds to the follower's response.
This separation enables us to construct a dataset that more accurately reflects the relationship between intended action and resulting state, thereby improving the quality of imitation learning.

However, two key issues remain.
First, generating both position and force commands increases the dimensionality of the model, making training more difficult. In many contact-rich tasks such as character writing, position trajectories are relatively easy to obtain, while force commands are often unknown. Thus, generating force commands from given position trajectories can reduce model complexity.
Second, learning-based motion generation generally suffers from low reproducibility compared to classical model-based methods. This can be mitigated by introducing feedback control, which improves robustness by correcting deviations during execution.

To address these issues, we create a hierarchical model extending the model proposed in~\cite{hayashi_ieeeaccess_2022}.
The upper layer handles the position trajectories to intelligently predict future states.
The output of the upper layer is hereafter referred to as the upper-layer trajectory.
The lower layer, on the other hand, predicts the state and command values for the next step based on the current state and the upper-layer trajectory.
We found that feedback control can fail when the lower layer has memory. 
To ensure controllability, we use a memoryless model such as a multilayer perceptron (MLP). 
Although MLPs alone have limited ability to generate complex motions, combining them with a memory-equipped upper layer enables intelligent motion generation.
Within the control loop, the error between the state predicted by the lower layer and the upper-layer trajectory is compensated for via proportional-integral-derivative (PID) control.
This allows the model to generate motion according to the upper-layer trajectory.
Furthermore, by implementing bilateral control-based imitation learning that generates force commands as the lower layer, it is possible to generate force commands that are appropriate for the desired upper-layer trajectory.
Therefore, when combined with path planning models \cite{planningtamizi2023review}, direct teaching \cite{ablettMultimodalForceMatchedImitation2023},
VLA models \cite{GR00TN1}\cite{octo_2023}\cite{kim24openvla}, 
etc., and given an intelligently predicted upper-layer trajectory, robot control with force commands becomes possible.
For this reason, we call the proposed method force generative imitation learning.

The contributions of this study are summarized as follows:
\begin{enumerate}
    \item We propose a hierarchical force generative imitation learning framework that explicitly separates a memory-based upper layer from a memoryless, feedback-controllable lower layer.
    \item The proposed framework enables contact-rich manipulation with nontrivial or ill-defined control objectives, which are difficult to address using conventional impedance or hybrid control, while still allowing stability to be ensured through classical feedback control.
    \item We reveal that embedding memory inside the feedback control loop can hinder stability, and demonstrate that time-scale separation allows learning-based force generation to achieve both robustness and stability, as validated in real robot writing experiments.
\end{enumerate}
In this study, a character writing task was conducted to validate the effectiveness of the proposed method. In the writing task, the NN was trained to perform various actions, and the generated motions were evaluated.
As a result of the experiments, as shown in Fig.~\ref{fig:Character_writing_tasks_improved_in_accuracy_by_the_proposed_method}, it was confirmed that the model, which had learned a variety of actions, was able to achieve both improved accuracy through PID control and stable contact through force control.

\section{Related Work}

\subsection{Bilateral Control-Based Imitation Learning}
Bilateral control is a type of teleoperation that synchronizes leader and follower robots while transmitting environmental reaction forces from the follower to the operator via the leader. 
This allows the operator to perform tasks using force-based skills.

From the follower's perspective, the leader's response serves as the command.
Thus, the leader's response can be treated as the action \ve{a} and the follower's response as the state \ve{s}, enabling the collection of action-state pairs (\ve{a}, \ve{s}) for supervised learning.

In conventional imitation learning, for example, an image may be used as the state \ve{s} and the next-step joint angles as the action \ve{a} \cite{10295965}\cite{fu2024mobile}. 
This assumes that the commanded and actual joint angles are identical, which typically requires slow movements to ensure that tracking errors remain negligible.
In contrast, bilateral control inherently accounts for command-response discrepancies, allowing data collection under non-ideal control conditions and enabling learning of high-speed motions \cite{Inami05062025}.

\subsection{Hierarchical Model}
Recently, there has been a growing interest in models inspired by the dual-process theory of human cognition, namely System 1 (fast, intuitive) and System 2 (slow, deliberative) thinking, and their effectiveness has been empirically validated in systems such as Figure's Helix and NVIDIA's GR00T N1 \cite{GR00TN1}.
These hierarchical models operate with different inference frequencies across layers: the lower layer, corresponding to System 1, performs fast, reactive inference at a high sampling rate, while the upper layer, analogous to System 2, performs slower, long-context inference with a lower sampling frequency.
However, a critical limitation shared by these approaches is that the upper and lower modules cannot be trained independently; optimizing one inherently depends on the other, complicating modularity and scalability.

In contrast, Hayashi $et~al.$ proposed a hierarchical model explicitly designed to enable independent training of the upper- and lower-level modules \cite{hayashi_ieeeaccess_2022}.
This not only simplifies the training process but also opens up the possibility of leveraging the upper-level module as a pretrained, task-agnostic policy module, while fine-tuning only the lower-level controller for task-specific adaptation. This modularity offers a clear advantage in terms of sample efficiency and generalization across tasks.

Moreover, many of these recent models are limited in that they only generate position commands and are not capable of producing force commands required for compliant and contact-rich manipulation. 
Our model addresses this shortcoming by incorporating force generation capabilities, thereby extending its applicability to a broader class of real-world robotic tasks where precise force interaction is essential.

\subsection{World Model with Control System}
Integrating control with NNs has been studied extensively using world models \cite{NIPS2015_a1afc58c}\cite{ha2018worldmodels}\cite{9577560}. 
World models are NNs that learn the structure of the environment from observational data and represent it in a latent space. By incorporating a mechanism to control errors in this latent space, it becomes possible to combine NNs with a control system. 

Although the model showed a high success rate in the plug insertion task \cite{9981610}, generalizing to diverse behaviors remains challenging.
We found that it is necessary to separate the model responsible for memory from the controllable part without memory.
Furthermore, mapping to latent space is complex; therefore, we employed simpler MLP models in this study.

\begin{figure}[!t]
    \centering
    \includegraphics[width=\linewidth]{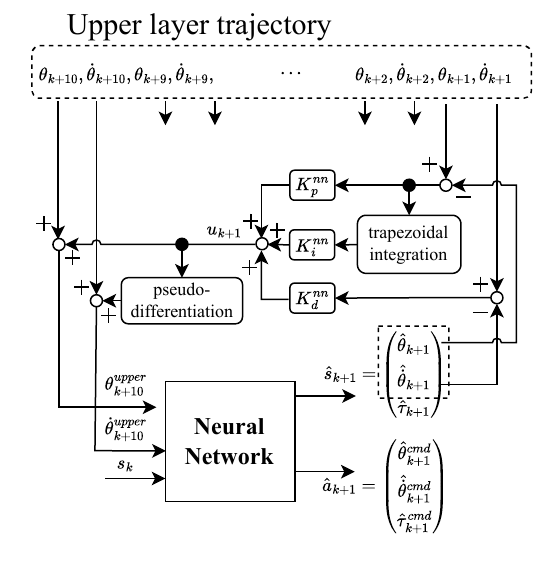}
    \caption{The proposed hierarchical model for correcting output errors}
    \label{fig:Proposed_PID_control}
\end{figure}

\section{Force Generative Imitation Learning}

\subsection{Hierarchical Architecture Separating Memory-Based and Memoryless Neural Networks}
In this study, a hierarchical model was employed to separate a model with memory from a model without memory.
The NN shown in Fig.~\ref{fig:Proposed_PID_control} corresponds to the lower layer, and the dotted line at the top corresponds to the upper layer.
The upper layer outputs angles $\bm{\theta}$ and angular velocities $\dot{\bm{\theta}}$ in 1-step increments from 1 step ahead to 10 steps ahead.
The outputs of the upper layer are updated in 10-step increments. 
While the extent to which the higher-level model should predict into the future depends on the task, it is difficult to determine an appropriate value. Therefore, in this study, we followed previous approaches and set it to 10 \cite{hayashi_ieeeaccess_2022}.

The [$\bm{\theta}_{k+10}$, $\dot{\bm{\theta}}_{k+10}$] given to the lower layer remains constant until the next update, and the output used in the control is the corresponding [$\bm{\theta}$, $\dot{\bm{\theta}}$] at each step.
Here, $k$ denotes the time step.
The model must predict future states and handle motions with memory.
In this study, the angles and angular velocities of the upper layer are given as the upper-layer trajectory.

On the other hand, the lower layer takes the upper-layer trajectory $[\bm{\theta}_{k+10}, \dot{\bm{\theta}}_{k+10}]$ 10 steps ahead and the current state ${\bm{s}_k}$ (joint angles, angular velocities, and torque) as input, and the state ${\hat{\bm{s}}_{k+1}}$ of the next step and the action ${\hat{{\bm{a}}}_{k+1}}$ are generated.
This does not require memory,
because it predicts values that interpolate between the desired future trajectory and the current state.
This structure eliminates the need for the lower layer to maintain internal states that make control difficult.
Because $\hat{\bm{a}}_{k+1}$ includes the 
force commands, the model can generate force commands from the position trajectory.
During autonomous execution, the ground-truth value of $\bm{a}_{k}$ is not available as input, which may cause errors to accumulate over time. However, a method to compensate for this issue has already been proposed in \cite{akagawa_jia_2023}.

In this study, upper-layer trajectories were collected by direct teaching.
This allows the robot to act in such a way that it reproduces the actions intuitively collected by the human.
In addition, since the upper layer is the angular trajectory of the robot, the proposed method can also be integrated with a model that intelligently predicts the robot's trajectory such as VLA models \cite{kim24openvla}.

\subsection{PID control}
This section describes the PID control introduced into the NN.
A conceptual diagram is shown in Fig.~\ref{fig:Proposed_PID_control}.
The NN predicts the state $\hat{\bm{s}}_{k+1}$ and the action $\hat{\bm{a}}_{k+1}$.
The $\hat{\bm{a}}_{k+1}$ is used as a command to the robot.
The angles $\hat{\bm{\theta}}_{k+1}$ and angular velocities $\hat{\dot{\bm{\theta}}}_{k+1}$ of state $\hat{\bm{s}}_{k+1}$ predicted by the NN are used for control.
Using $\bm{\theta}_{k+1}$ and $\dot{\bm{\theta}}_{k+1}$ of the upper-layer trajectory, the PID control error $\bm{{u}}_{k+1}$ is calculated as follows:
\begin{equation}
    \begin{split}
        \bm{u}_{k+1} = 
        &{K_p^{nn}}(\bm{\theta}_{k+1} - \hat{\bm{\theta}}_{k+1}) \\
        + &{K_d^{nn}}(\bm{\dot\theta}_{k+1} - \hat{\bm{\dot\theta}}_{k+1}) \\
        + &{K_i^{nn}}\int_{0}^{k+1}(\bm{\theta}_{t+1} - \hat{\bm{\theta}}_{t+1})dt
    \end{split}
\end{equation}
where $K_p^{nn}$, $K_d^{nn}$, and $K_i^{nn}$ are the proportional, derivative, and integral gains in the control.
In addition, the gains are identical across all axes.
The integration was performed using the trapezoidal rule.
Then, $\bm{u}_{k+1}$ was differentiated with respect to time to compute $\dot{\bm{u}}_{k+1}$. A pseudo-differentiation filter with a cutoff frequency of 0.3~Hz was used. The calculated $\bm{u}_{k+1}$ and $\dot{\bm{u}}_{k+1}$ were added to the upper-layer angles $\bm{\theta}_{ k+10}$ and angular velocities $\dot{\bm{\theta}}_{k+10}$.
Finally, this was input into the NN as the angle $\bm{\theta}^{upper}_{k+10}$ and angular velocity $\dot{\bm{\theta}}^{upper}_{k+10}$ from the upper layer. This model is expected to generate command and response values based on the current state and the upper-layer trajectory, which includes error correction terms. 
As described above, the proposed method can generate appropriate force commands considering trajectory prediction errors.

\begin{figure*}[!t]
    \centering
    \begin{minipage}{0.3\linewidth}
        \centering
        \includegraphics[width=\linewidth]{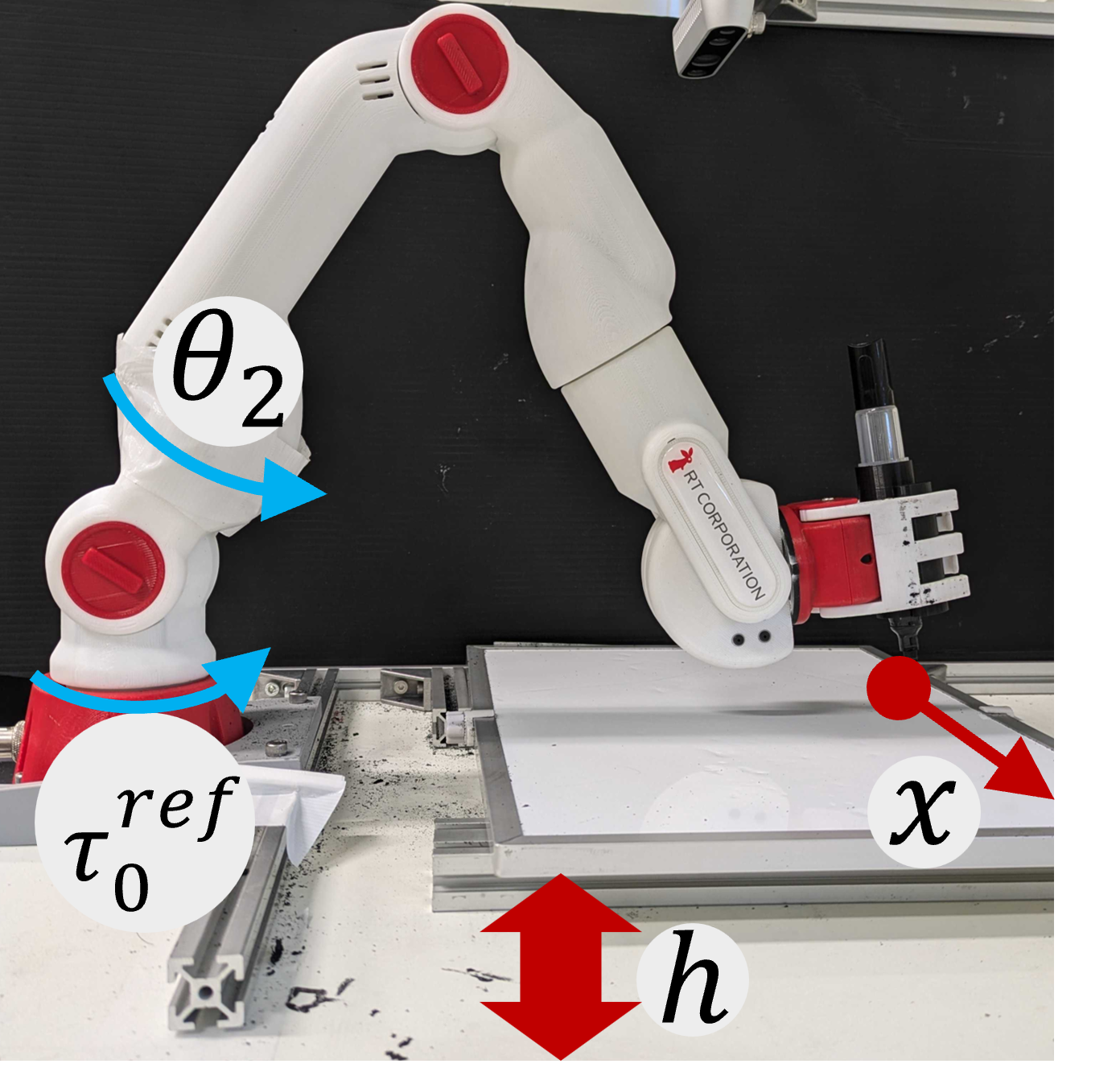}
        \caption{Task environment}
        \label{fig:task_overview}
    \end{minipage}
    \hfill
    \begin{minipage}{0.35\linewidth}
        \centering
        \includegraphics[width=\linewidth]{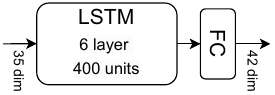}
        \caption{LSTM model}
        \label{fig:LSTM_model}
        \vspace{0.5em}
        \includegraphics[width=\linewidth]{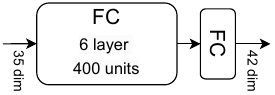}
        \caption{MLP model}
        \label{fig:MLP_model}
    \end{minipage}
    \hfill
    \begin{minipage}{0.3\linewidth}
        \centering
        \includegraphics[width=\linewidth]{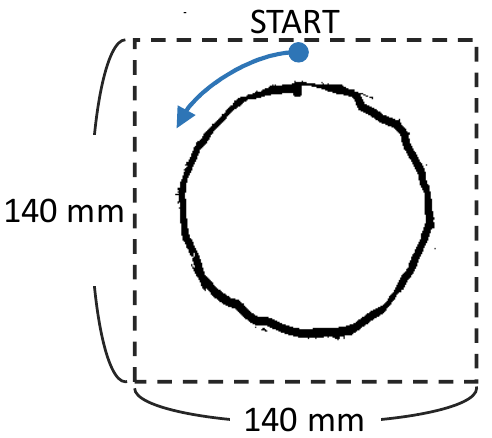}
        \caption{Circle diagram collected via direct teaching}
        \label{fig:Circle_diagram_collected_by_direct_teaching}
    \end{minipage}
\end{figure*}

\section{Experiment}
\subsection{Manipulator}
In this study, CRANE-X7 manipulators (RT Corp., Tokyo) were employed.
Each manipulator has seven degrees of freedom, and the gripper has one degree of freedom.
The gripper was replaced with a cross-structured hand \cite{yamane2023soft}.
Since controlling a manipulator with seven degrees of freedom can be challenging for humans, joint 2, as shown in Fig.~\ref{fig:task_overview}, was fixed using position control, effectively reducing the system to a six-degree-of-freedom manipulator.
A position/force hybrid controller was used to control each joint with a control frequency of 500 Hz.
This robot was equipped only with joint angle sensors; thus, angular velocities were obtained using pseudo-differentiation.
Furthermore, torques were measured without torque sensors by employing a reaction force observer \cite{RTOB}.
Note that the robot control method and control parameters are identical to those in \cite{saigusa_ieeeaccess_2022}.

\subsection{Task design and data collection}
In the experiment, a character writing task was performed to demonstrate the effectiveness of the proposed method.
The proposed method is not limited to any specific task; however, writing was chosen as a contact-rich task in which the correctness of motion trajectories can be readily evaluated.
The experimental environment is shown in Fig.~\ref{fig:task_overview}. The task was performed with the robot holding the pen from the beginning.
The robot's writing was recorded using an Intel RealSense D435i placed above a whiteboard.

Training and validation data were collected from five different lines and two different circular trajectories using bilateral control, as shown at the top of Fig.~\ref{fig:Character_writing_tasks_improved_in_accuracy_by_the_proposed_method}.
To improve generalization across height, data were collected at whiteboard heights of 0 cm and 2 cm, as indicated by $h$ in Fig.~\ref{fig:task_overview}.
Motion sequences in five trials were collected for each of the five types of lines and two types of circles at two different heights, for a total of 70 trials.
56 trials of the collected data were used as training data and 14 trials were used as validation data.

It should be noted that, while joint angles, angular velocities, and torques are available during training, only joint angles are provided during autonomous operation.
Therefore, it is necessary to estimate torques solely from the available angle data.

\subsection{Training NN}
The NN model used in this study is shown in Fig.~\ref{fig:LSTM_model} and Fig.~\ref{fig:MLP_model}.
The inputs to the lower layer consisted of five variables: the current joint angles, angular velocities, and torque responses, as well as the joint angle and angular velocity commands at 10 steps ahead provided by the upper layer.
Since each of these quantities was defined for seven degrees of freedom, six joints (joint 1 to joint 7, excluding joint 2) plus the hand, the total input dimension was $5$ (variables) $\times 7$ (DOF)$= 35$,
while the outputs consisted of six variables: the joint angles, angular velocities, and torque responses at the next step, together with the corresponding command values of joint angles, angular velocities, and torques. Therefore, the total output dimension was $6$ (variables) $\times 7$ (DOF)$= 42$.
Fig.~\ref{fig:LSTM_model} shows the LSTM model with internal state and Fig.~\ref{fig:MLP_model} shows the proposed MLP model without internal state.
The LSTM consists of 6 layers, each with 400 dimensions, and one fully connected layer, for a total of 7 layers.
The MLP consists of seven fully connected layers, each with 400 dimensions, and uses the tanh function as the activation function in all layers except the last one. 
Note that the NN was trained on the collected training data and the PID controller was not used during training.

For the input data, zero-mean Gaussian noise with a variance of 0.01 was added to each collected data sequence. The addition of this noise allows for diversity in the data and at the same time, allows for the construction of a model that is robust to noise.
A zero-phase low-pass filter with a cutoff frequency of 20 rad/s was applied to the training data. Applying this filter helps prevent the neural network from overfitting to high-frequency noise.
In addition, the input and training data, which were at 2 ms intervals, were resampled every 10 steps by shifting the starting point so that the dataset was at 20 ms intervals. 
This process increased the data volume by a factor of 10 \cite{rahmatizadeh_2018}.
In addition, a padding process was used to copy the last value to make the length of each sequence uniform.

During training, the data were normalized to have a mean of 0 and a standard deviation of 1.
The mean squared error (MSE) was used as the loss function and Adam \cite{KingBa15} was used as the optimization method.
The learning rate was set to 0.0001, the batch size was set to 128, and training was performed for up to 5000 epochs.
The input window length corresponds to the length of the data collected in each trial. When the sequence lengths differ, they are unified by padding through duplication of the final value. Gradient clipping was not used.
Among these, the model at epoch 500 with the lowest validation loss was used in the experiment.
The validation loss decreased until around 500 epochs and increased thereafter, suggesting overfitting. Although models trained beyond 500 epochs were not evaluated, further training may reduce trajectory generation performance under unseen conditions.

\subsection{Verification of PID Gain Characteristics}
In this section, we compared the responses of the robot when the gains in the proposed method of PID control were changed.
The task was to write a circular trajectory shown in Fig.~\ref{fig:Circle_diagram_collected_by_direct_teaching} obtained by direct teaching.
The figure is a binary representation of the drawn trajectory, and the motion was to make contact at the starting point at the top and move counterclockwise.

To evaluate task performance, we compared the pen nib's Cartesian x-coordinate trajectory between the upper-layer output and the robot's autonomous motion. 
The origin of the x-coordinate was defined as the pen's position at the start of the robot's action, as shown in Fig.~\ref{fig:task_overview}.  

\begin{figure*}[!t]
    \centering
    \begin{minipage}[t]{0.32\textwidth}
        \centering
        \includegraphics[width=\linewidth]{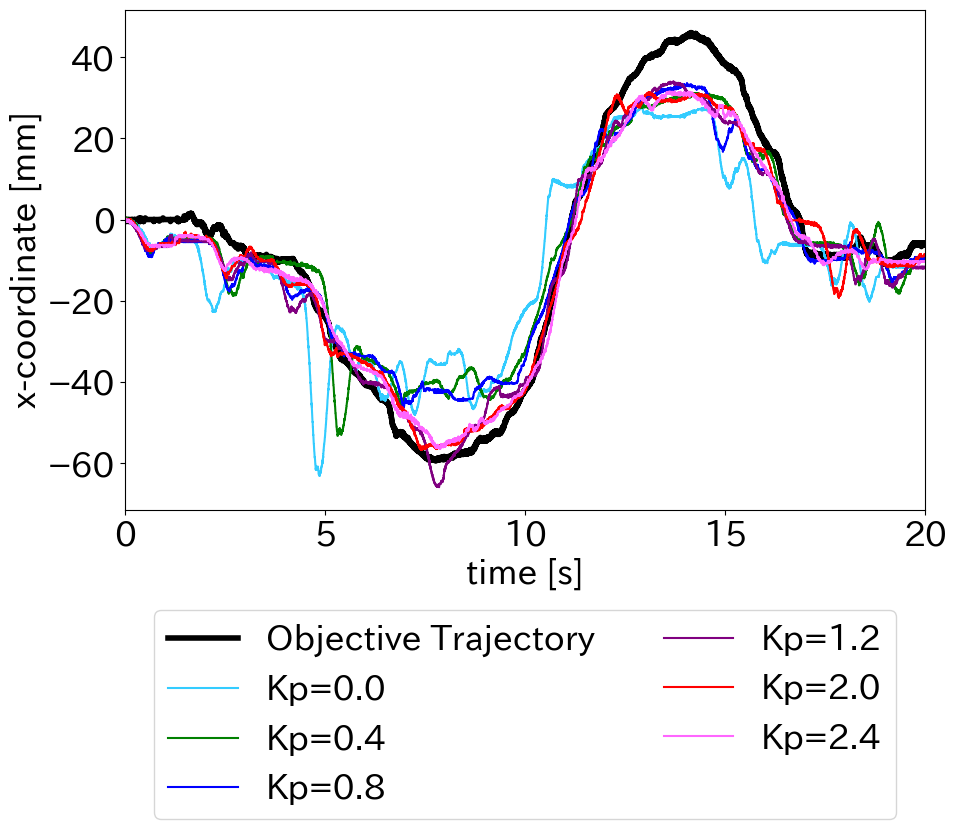}
        \caption{Pen nib x-coordinate with various $K_p$}
        \label{fig:Kp_x_comparison}
    \end{minipage}
    \hfill
    \begin{minipage}[t]{0.32\textwidth}
        \centering
        \includegraphics[width=\linewidth]{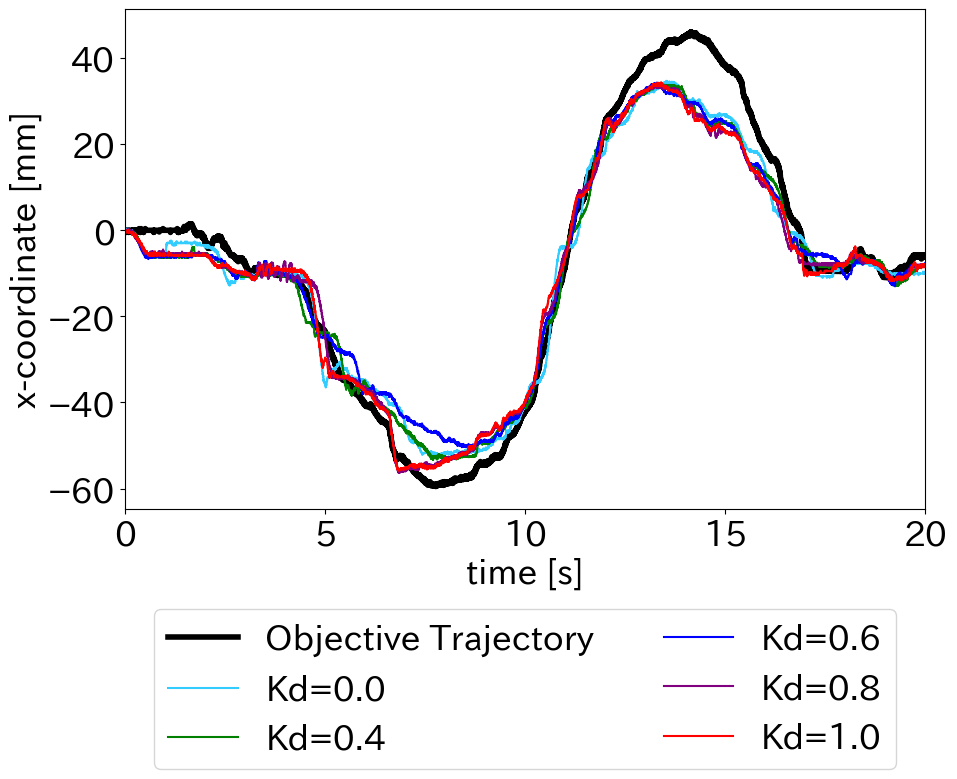}
        \caption{Pen nib x-coordinate with various $K_d$}
        \label{fig:Kd_x_comparison}
    \end{minipage}
    \hfill
    \begin{minipage}[t]{0.32\textwidth}
        \centering
        \includegraphics[width=\linewidth]{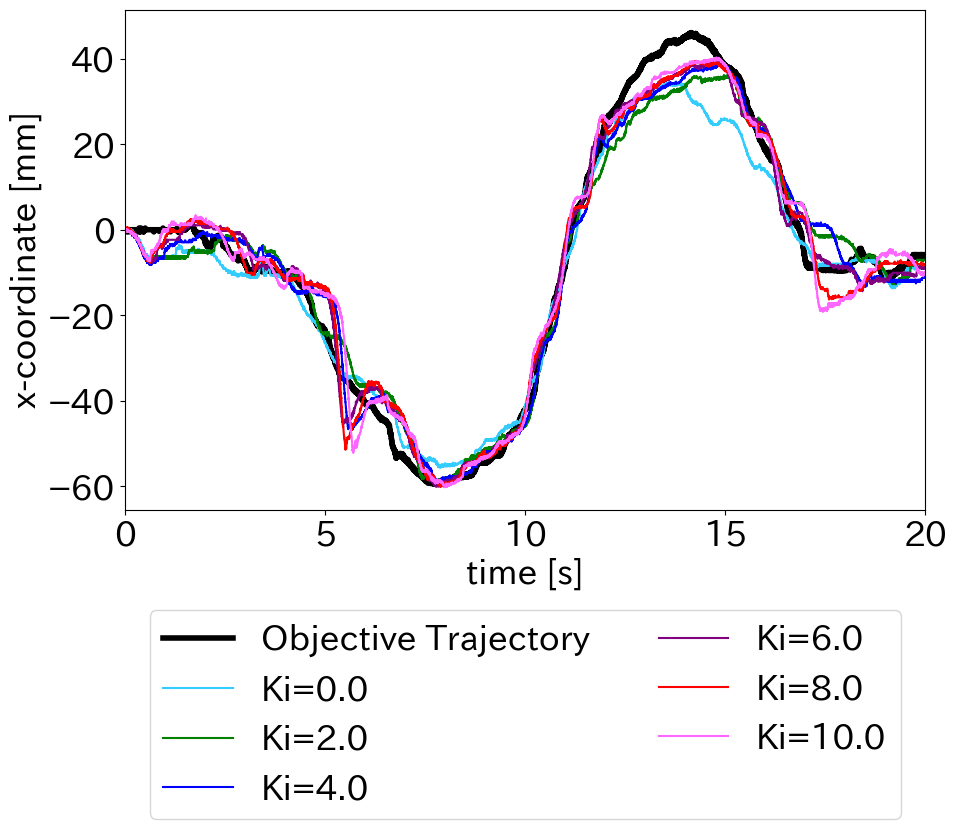}
        \caption{Pen nib x-coordinate with various $K_i$}
        \label{fig:Ki_x_comparison}
    \end{minipage}
\end{figure*}

\begin{figure*}[t]
  \centering
  \subfloat[Comparison with different $K_p$]{
    \includegraphics[width=0.31\linewidth]{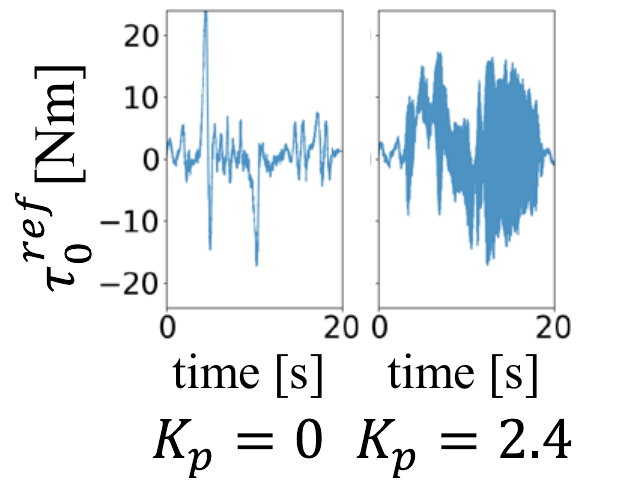}
    \label{fig:compare_torque_Kp}
  }
  \hfill
  \subfloat[Comparison with different $K_d$]{
    \includegraphics[width=0.31\linewidth]{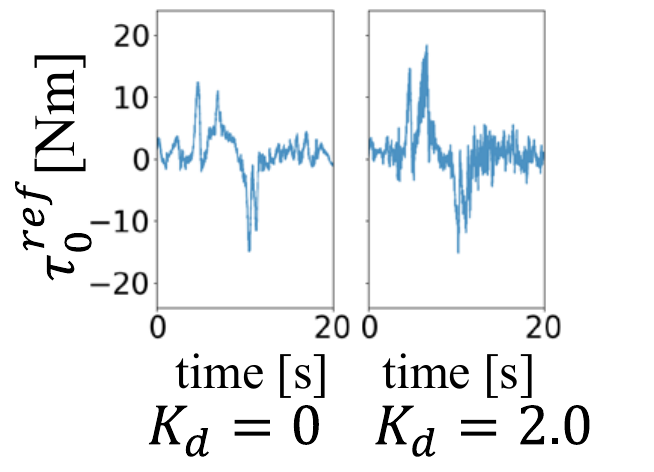}
    \label{fig:compare_torque_Kd}
  }
  \hfill
  \subfloat[Comparison with different $K_i$]{
    \includegraphics[width=0.31\linewidth]{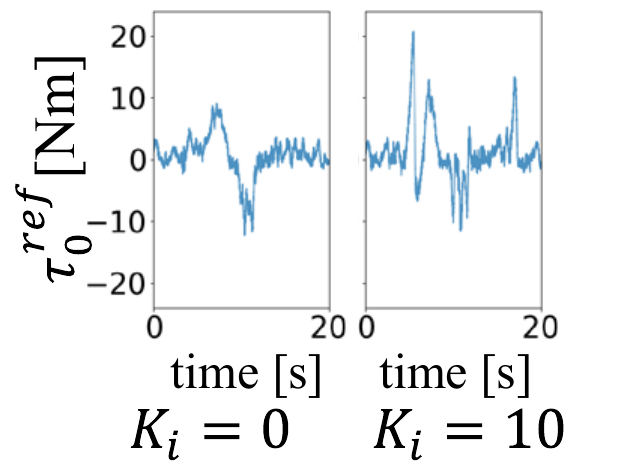}
    \label{fig:compare_torque_Ki}
  }
  \caption{Comparison of torque reference with different PID gains.}
    \label{fig:Comparison_Chart_of_PID_Gains_and_Torque_Reference_Values}
\end{figure*}

Comparisons of behavior with changes in $K_p^{nn}$ were made by increasing $K_p^{nn}$ from 0.0 to 2.4 in increments of 0.4.
Here, $K_d^{nn}$ and $K_i^{nn}$ were fixed at 0.
Fig.~\ref{fig:Kp_x_comparison} shows the trajectories of the x-coordinate of the pen nib during autonomous operation.
The solid black lines indicate the upper-layer trajectories.
Fig.~\ref{fig:Comparison_Chart_of_PID_Gains_and_Torque_Reference_Values}~(a) shows the torque reference values $\bm{\tau}^{ref}_0$ at the base joint, as shown in Fig.~\ref{fig:task_overview}, when $K_d^{nn}$ was 0.0 and 2.4.
Fig.~\ref{fig:Kp_x_comparison} shows that when $K_p^{nn}$ was set to 0, the x-coordinate exhibited oscillatory behavior compared to the upper-layer trajectory.
On the other hand, when the gain of $K_p^{nn}$ was increased, the oscillatory behavior was suppressed and the motion followed the upper-layer circular trajectory.
These results indicate that the introduction of P control contributed to the generation of commands that allow the NN to follow the upper-layer trajectory.
However, as shown in Fig.~\ref{fig:Comparison_Chart_of_PID_Gains_and_Torque_Reference_Values}~(a), oscillations and increases in the torque reference value were observed as the gain increased.
Although this did not lead to problems in task execution in this experiment, the stability and safety of the robot may be lost as the gain increases.
Therefore, task performance and robot stability should be evaluated to set appropriate gains in the future.

Comparisons of the behavior of the robot with changes in $K_d^{nn}$ were made by increasing $K_d^{nn}$ from 0.0 to 1.0 in 0.2 increments.
Here, $K_p^{nn}$ and $K_i^{nn}$ were fixed at 1.6 and 0, respectively.
Fig.~\ref{fig:Kd_x_comparison} shows the trajectories of the x-coordinate of the pen nib during autonomous operation.
Fig.~\ref{fig:Comparison_Chart_of_PID_Gains_and_Torque_Reference_Values}~(b) shows the torque reference values $\bm{\tau}^{ref}_0$ at the base joint, as shown in Fig.~\ref{fig:task_overview}, when $K_d^{nn}$ was 0.0, and 1.0.
In Fig.~\ref{fig:Kd_x_comparison}, no significant change in behavior was observed as a result of increasing $K_d^{nn}$.
However, local oscillations were observed in the x-coordinate when $K_d^{nn}$ was 0.8 or 1.0.
Oscillations and increases in the torque reference were also observed when $K_d^{nn}$ was increased.
In this study, the output of the PID controller was converted to the angular velocity dimension by performing pseudo differentiation.
The derivative calculation was performed in continuous time, which seemed to induce computational errors that negatively impacted the NN.
Although some negative effects were observed with D control in this study, changes in the torque reference values were observed when $K_d^{nn}$ was changed, suggesting that the introduction of D control with appropriate calculation methods will have a positive effect on the system in the future.

Comparisons of behavior with changes in $K_i^{nn}$ were made by increasing $K_i^{nn}$ from 0.0 to 10.0 in increments of 2.0.
Here, $K_p^{nn}$ and $K_d^{nn}$ were fixed at 1.6 and 0.2, respectively.
The trajectories of the x-coordinate of the pen for different $K_i^{nn}$ gains are shown in Fig.~\ref{fig:Ki_x_comparison}.
Fig.~\ref{fig:Comparison_Chart_of_PID_Gains_and_Torque_Reference_Values}~(c) shows extracts of the torque reference values $\bm{\tau}^{ref}_0$ at the base joint, as shown in Fig.~\ref{fig:task_overview}, when $K_i^{nn}$ was 0.0 and 10.0.
Fig.~\ref{fig:Ki_x_comparison} shows that as the $K_i^{nn}$ gain was increased, the deviation was reduced near the top of the x-coordinate waveform.
However, near the bottom of the x-coordinate waveform, previously unseen waveform peaks appeared.
On the other hand, as shown in Fig.~\ref{fig:Comparison_Chart_of_PID_Gains_and_Torque_Reference_Values}~(c), even as \(K_i^{nn}\) increased, there was no excessive increase in the torque reference value, but there were changes in the waveform.
These results indicate that the introduction of I control affects the generation of NN reference values.
However, there were cases where this method had a negative effect.
Since the operating period of NN was large, it was considered to be due to the effect of errors as in D control.

The results show that the generation of NN command values varied with changes in gain.
Therefore, the appropriate introduction of PID control can improve the accuracy of the task.
However, the method did not use discrete-time implementation for differential and integral calculations, which could lead to calculation errors.
Therefore, as future work, it is necessary to introduce discrete control techniques to account for calculation errors in PID control.

\begin{figure}[!t]
    \centering
        \centering
        \includegraphics[width=60mm]{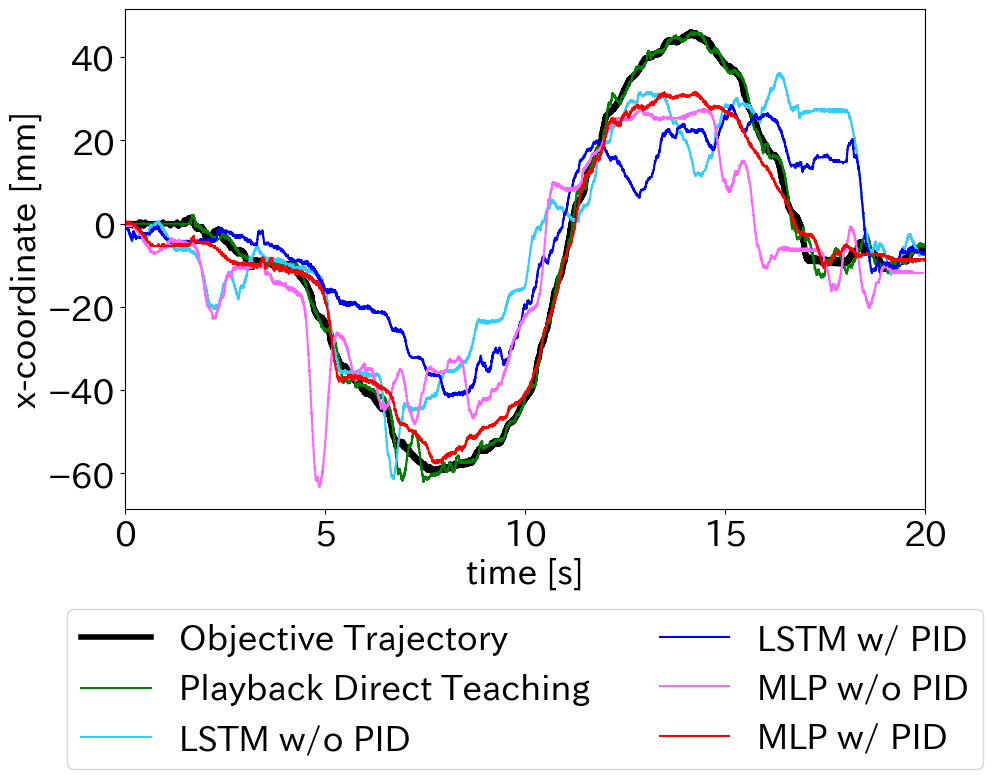}
        \caption{Pen nib x-coordinate with various models}
        \label{fig:ordinal_model_x_comparison}

\end{figure}

\subsection{Comparison with conventional methods}
In this section, we compared autonomous operations using response values collected during direct teaching as command values, the traditional LSTM model, and our proposed MLP model. 

After collecting movements via direct teaching, it seems straightforward to use the collected response values as command values to reproduce the collected movements. 
However, this method faces difficulties in performing contact tasks due to the lack of force control.
Here, we compared autonomous operations using response values from direct teaching (referred to as ``Playback Direct Teaching") with our proposed method. In addition, we included LSTMs as a traditional model for handling time series data in our comparison. While LSTMs are known to generate movements based on past memories, the strong influence of internal states makes it difficult to integrate control with NNs. Therefore, we compared autonomous operations with and without control for the LSTM model and also performed similar operations with the MLP model.
The gains were adjusted in an exploratory manner, and the following experiments were conducted with: \( K_p^{nn} = 2.0 \), \( K_i^{nn} = 0.0 \), and \( K_d^{nn} = 0.2 \).

In the experiments, we collected movements for drawing the circle, characters A and B via direct teaching, using these as upper-layer trajectories. As a comparison metric, we calculated the Intersection over Union (IoU) between the upper-layer and the actual drawn shapes. For the circle movements, we compared the trajectories of the pen's x-coordinate as in the previous section.

Fig.~\ref{fig:ordinal_model_x_comparison} shows the $x$ trajectories of the pen nib for five models during the circle drawing task.
Drawings of the circle, A and B are shown in Fig.~\ref{fig:Character_writing_tasks_improved_in_accuracy_by_the_proposed_method} and IoU values are given below.
The actual drawings are shown in black, and the upper-layer trajectories are shown in cyan, with the degree of agreement illustrated by superimposing them. 
Fig.~\ref{fig:ordinal_model_x_comparison} indicates that ``Playback Direct Teaching'' followed the intended trajectory.
However, as shown in Fig.~\ref{fig:Character_writing_tasks_improved_in_accuracy_by_the_proposed_method}, the drawings of the circle, A and B, were not accurate. 
This can be attributed to the lack of force control in the robot, making it difficult to press against the board surface during operation. This suggests that autonomous operation using only angle commands is challenging for tasks involving writing characters that require contact.

\begin{table*}[!t]
    \centering
    \caption{IoU results comparing MLP w/o PID and MLP w/ PID}
    \label{fig:IoU_results}
    \renewcommand{\arraystretch}{1.5}  
    \scalebox{1.2}{
    \begin{tabular}{cc|ccc||cc}
        \hline
        \textbf{Model} & \textbf{Height} & \textbf{Character 2} & \textbf{Character 3} & \textbf{Character ABCD} & \textbf{Average} & \textbf{Total Value} \\
        \hline
        \multirow{3}{*}{\rotatebox{-90}{\begin{minipage}[c][1cm][c]{2cm}  MLP \\ w/o PID\end{minipage}}} & 0cm & $0.097 \pm 0.036$ & \bm{$0.182 \pm 0.048$} & $0.122 \pm 0.009$ & $0.134 \pm 0.031$ &\\
        & 1cm & $0.110 \pm 0.060$ & $0.059 \pm 0.010$ & $0.165 \pm 0.028$ & $0.111 \pm 0.033$ & $0.121 \pm 0.041$ \\
        & 2cm & $0.106 \pm 0.045$ & $0.097 \pm 0.066$ & $0.150 \pm 0.071$ & $0.118 \pm 0.061$ & \\
        \hline
        \multirow{3}{*}{\rotatebox{-90}{\begin{minipage}[c][1cm][c]{2cm}  MLP \\ w/ PID\end{minipage}}} & 0cm & $\bm{0.108 \pm 0.020}$ & $\textcolor{black}{{0.112 \pm 0.035}}$ & $\bm{0.218 \pm 0.014}$ & $\bm{0.146 \pm 0.023}$ &  \\
        & 1cm & $\bm{0.222 \pm 0.032}$ & $\bm{0.173 \pm 0.021}$ & $\bm{0.230 \pm 0.009}$ & $\bm{0.270 \pm 0.047}$ & $\textcolor{black}{\bm{0.208 \pm 0.030}}$ \\
        & 2cm & $\bm{0.212 \pm 0.055}$ & $\bm{0.332 \pm 0.055}$ & $\bm{0.266 \pm 0.030}$ & $\bm{0.208 \pm 0.021}$ &  \\
        \hline
    \end{tabular}
    }
\end{table*}

\begin{figure}[!t]
    \centering
    \includegraphics[width=70mm]{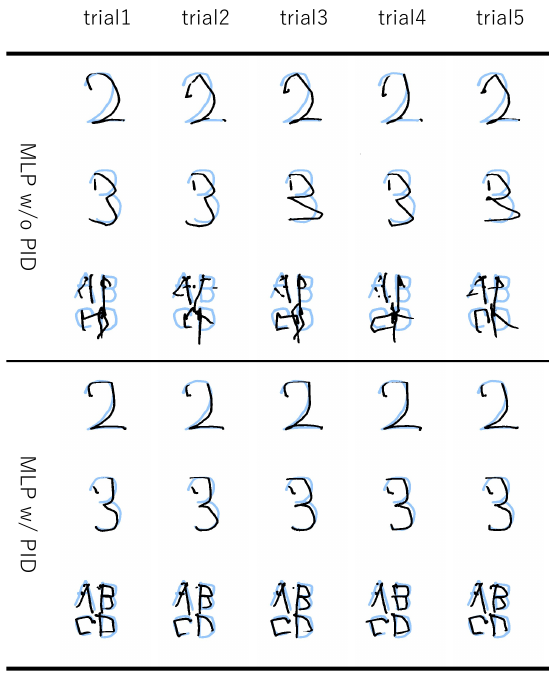}
    \caption{Characters drawn by MLP w/o PID and MLP w/ PID when the whiteboard height is set at 1 cm}
    \label{fig:1cm_results}
\end{figure}

While force generative imitation learning has the advantage of inferring force commands according to the current force response, it also modifies the position commands.
In other words, it generates commands that trade off between position and force commands, and therefore does not strictly follow the human-taught position trajectory.
Then, the conventional LSTM model was not able to follow the upper-layer trajectory as shown in Fig.~\ref{fig:ordinal_model_x_comparison}.
Introducing PID control to the LSTM did not improve its accuracy.
This can be seen in Fig.~\ref{fig:Character_writing_tasks_improved_in_accuracy_by_the_proposed_method}.
For the MLP model without PID control, tracking to the upper-layer trajectory was insufficient, as shown in Fig.~\ref{fig:ordinal_model_x_comparison}. However, by incorporating PID control into the proposed method, the tracking performance was improved. Fig.~\ref{fig:Character_writing_tasks_improved_in_accuracy_by_the_proposed_method}, which shows the circle, A and B, also shows that the autonomous operation of the MLP with PID control was the closest to the intended image, with the highest IoU values for the characters except for character B.
For character B, the IoU was small even though the shape was closest to the intended B among the tested models.
This may be because the IoU calculation evaluated the degree of overlap and could not evaluate the accuracy of the shape.
In addition, it was difficult to draw the right half of the circular trajectory. 
This seemed to be caused by insufficient movement from bottom to top in the training data.
Consequently, even with the introduction of the control, it was difficult to reproduce the bottom-to-top circular motion.

These results show that by incorporating control, MLP can generate movements closer to the intended trajectories based on learned behaviors. This was possible because MLP has no internal state, allowing it to generate appropriate command values in response to feedback from the PID controller.

\subsection{Verification of generalizability}
In this experiment, we tested the generalization performance of models without and with control in the learned MLP model.  
We collected characters ``2", ``3" and the string ``ABCD" by direct teaching and the characters and string were used as upper-layer trajectories.  
The task labeled ``ABCD" consists of writing the sequence ABCD four times in a row and contains the smallest characters used in any of the previous writing tasks.
In the validation, the experiment was conducted with the whiteboard and desk heights shown in Fig.~\ref{fig:task_overview} set to \( h = 0\ \mathrm{cm} \), \( h = 1\ \mathrm{cm} \), and \( h = 2\ \mathrm{cm} \) where \( 0\ \mathrm{cm} \) and \( 2\ \mathrm{cm} \) were the height used during learning and \( 1\ \mathrm{cm} \) was the unlearned height. 
The autonomous operations were performed five times for each model. 
They were evaluated by calculating the mean and variance of the IoU between the drawn characters and the upper-layer trajectory.

The table summarizing the IoU results is presented in Table~\ref{fig:IoU_results}. In addition, the actual drawings of 1~cm are shown in Fig.~\ref{fig:1cm_results}. In this figure, the actual trajectories drawn are shown in black and the upper-layer trajectories are shown in cyan.
The IoU results confirmed that the introduction of PID control improved the IoU values for many characters and heights, a trend that was also reflected in the Total Value. 
At the untrained height of 1~cm, the ability to write characters was maintained with and without PID control, but the precision of character writing was improved with PID control. 
These results are evident in Fig.~\ref{fig:1cm_results}, which shows a high degree of congruence in the characters drawn, particularly in the string ``ABCD", where tracking of the upper-layer trajectory was improved.

Previously, the robot had difficulty performing the task of writing smaller characters based on learning larger characters. In particular, the ``draw a line and stop" action for smaller characters proved challenging, with many instances of overshooting the intended line when no control was applied. However, the integration of the proposed PID control in this study reduced the number of trials that deviated from the upper-layer trajectory, thus improving the accuracy of the characters.

Conversely, at a height of 0~cm, the IoU value for character 3 was higher without PID control. 
The low IoU value was due to incomplete matching with the upper-layer trajectory.
Even if the shape of the drawn character was improved by the control, the IoU value would not be high if it deviated slightly from the upper-layer trajectory.
As a result, the IoU values were small because shape evaluation was not possible.
It is also expected that the introduction of appropriate integral control and iterative learning control will correct any deviations that occur.

\section{Discussion}
\subsection{Stability and Memory Interaction}
Memory-based models inherently exploit past information, which can be interpreted as an implicit internal feedback mechanism. When combined with an explicit feedback controller, this results in coupled feedback loops, making stability guarantees difficult, particularly for nonlinear neural networks. Our hierarchical architecture mitigates this issue via time-scale separation: the memory-equipped upper layer operates at a much lower frequency than the lower-layer feedback controller. Consequently, from the lower layer's perspective, the upper-layer output is quasi-static and effectively acts as a feedforward command, preventing interference with the stability of the feedback control loop.

\subsection{Limitations}
Whether the lower layer can be implemented as a memoryless model may depend on the specific task.
At present, we have not conducted sufficient verification to determine how far this can be generalized. 
However, since the lower-level model generates the next-step action and state by interpolating between the current state and the action 10 steps ahead, we believe that assuming the lower level to be memoryless is not a significant issue, as long as the upper level can appropriately predict future commands.
While it is possible that the appropriateness of updating every 10 steps is also task-dependent, in our experience across various implemented tasks, using 10 steps has not caused any problems.
For more complex tasks, it might be worth considering making this step size itself a learnable parameter.

Furthermore, the current evaluation focuses on a planar writing task, where positions perpendicular to the surface do not need to be explicitly considered. This reduces the effective dimensionality of the control problem and likely facilitates stable feedback control. As a result, although the proposed method performs well in this setting, its robustness may not be guaranteed for more complex contact-rich tasks, such as writing or wiping on three-dimensional surfaces, which require full three-dimensional control.

\subsection{Comparison with Conventional Methods}
Numerous studies have been reported on enabling robots to perform writing tasks, employing approaches such as impedance control \cite{6523093} and hybrid control \cite{6224651}.
Through rigorous control system design, these methods have achieved industrial-level reproducibility of motions, and the technology can be considered to have already reached a mature stage.
However, the value of the proposed method lies in demonstrating that a framework for imitation learning, capable of reproducing general skillful motions beyond character writing tasks, can achieve performance comparable to rule-based feedback control.

\section{Conclusion}
In this study, we proposed force generative imitation learning with a PID controller to compensate for output errors of bilateral control-based imitation learning.
By collecting angular trajectory data through intuitive direct teaching, the proposed method successfully generated force commands.
In the performance evaluation by changing the PID gain, the command generation was confirmed to follow the upper-layer trajectory by adjusting the P-gain, and the accuracy of the task was improved.  
In addition, compared with the conventional operation with only position commands and the LSTM model with internal state, the autonomous operation with PID control incorporated into MLP without internal state showed the highest accuracy.  

On the other hand, the control of the proposed method needs further improvement.  
At this stage, the stability of the NN with PID control has not been fully evaluated, and the robot may take unintended actions due to the introduction of the controller.
Therefore, the optimization of gains that ensure the stability and controllability of the robot behavior is an issue.
We would also like to explore further possibilities for control by improving the computation to take into account discrete control and adopting other controls such as model predictive control.

\bibliographystyle{ieeetr}

\begin{thebibliography}{10}

\bibitem{survey_on_humanrobot}
V.~Villani, F.~Pini, F.~Leali, and C.~Secchi, ``Survey on human-robot
  collaboration in industrial settings: Safety, intuitive interfaces and
  applications,'' {\em Mechatronics}, vol.~55, pp.~248--266, 2018.

\bibitem{Industrial_Robot_Control_Trends}
J.~Arents and M.~Greitans, ``Smart industrial robot control trends, challenges
  and opportunities within manufacturing,'' {\em Applied Sciences}, vol.~12,
  no.~2, 2022.

\bibitem{weinberg2024survey}
A.~I. Weinberg, A.~Shirizly, O.~Azulay, and A.~Sintov, ``Survey of
  learning-based approaches for robotic in-hand manipulation,'' {\em Frontiers
  in Robotics and AI}, vol.~11, p.~1455431, 2024.

\bibitem{pmlr-v205-zhu23a}
Y.~Zhu, A.~Joshi, P.~Stone, and Y.~Zhu, ``Viola: Imitation learning for
  vision-based manipulation with object proposal priors,'' in {\em Conference
  on Robot Learning}, pp.~1199--1210, PMLR, 2023.

\bibitem{zhao2023learningfinegrainedbimanualmanipulation}
T.~Z. Zhao, V.~Kumar, S.~Levine, and C.~Finn, ``Learning fine-grained bimanual
  manipulation with low-cost hardware,'' 2023.

\bibitem{wang2023mimicplay}
C.~Wang, L.~Fan, J.~Sun, R.~Zhang, L.~Fei-Fei, D.~Xu, Y.~Zhu, and
  A.~Anandkumar, ``Mimicplay: Long-horizon imitation learning by watching human
  play,'' {\em arXiv preprint arXiv:2302.12422}, 2023.

\bibitem{fu2024mobile}
Z.~Fu, T.~Z. Zhao, and C.~Finn, ``Mobile aloha: Learning bimanual mobile
  manipulation with low-cost whole-body teleoperation,'' in {\em {Conference on
  Robot Learning (CoRL)}}, 2024.

\bibitem{octo_2023}
{Octo Model Team}, D.~Ghosh, H.~Walke, K.~Pertsch, K.~Black, O.~Mees,
  S.~Dasari, J.~Hejna, C.~Xu, J.~Luo, T.~Kreiman, Y.~Tan, L.~Y. Chen,
  P.~Sanketi, Q.~Vuong, T.~Xiao, D.~Sadigh, C.~Finn, and S.~Levine, ``Octo: An
  open-source generalist robot policy,'' in {\em Proceedings of Robotics:
  Science and Systems}, (Delft, Netherlands), 2024.

\bibitem{saigusa_ieeeaccess_2022}
Y.~Saigusa, S.~Sakaino, and T.~Tsuji, ``Imitation learning for nonprehensile
  manipulation through self-supervised learning considering motion speed,''
  {\em IEEE Access}, vol.~10, pp.~68291--68306, 2022.

\bibitem{akagawa_jia_2023}
T.~Akagawa and S.~Sakaino, ``Autoregressive model considering low frequency
  errors in command for bilateral control-based imitation learning,'' {\em IEEJ
  Journal of Industry Applications}, vol.~12, no.~1, pp.~26--32, 2023.

\bibitem{yamane2023soft}
K.~Yamane, Y.~Saigusa, S.~Sakaino, and T.~Tsuji, ``Soft and rigid object
  grasping with cross-structure hand using bilateral control-based imitation
  learning,'' {\em IEEE Robotics and Automation Letters}, vol.~9, no.~2,
  pp.~1198--1205, 2024.

\bibitem{buamanee2024bi}
T.~Buamanee, M.~Kobayashi, Y.~Uranishi, and H.~Takemura, ``Bi-act: Bilateral
  control-based imitation learning via action chunking with transformer,'' in
  {\em 2024 IEEE International Conference on Advanced Intelligent Mechatronics
  (AIM)}, pp.~410--415, IEEE, 2024.

\bibitem{Masato_Kobayashi202524004380}
M.~Kobayashi, T.~Buamanee, Y.~Uranishi, and H.~Takemura, ``Ilbit: Imitation
  learning for robot using position and torque information based on bilateral
  control with transformer,'' {\em IEEJ Journal of Industry Applications},
  vol.~14, no.~2, pp.~161--168, 2025.

\bibitem{Inami05062025}
K.~Inami, M.~Konosu, K.~Yamane, N.~Masuya, Y.~Li, Y.-H. Shu, H.~Sato, S.~Homma,
  and S.~Sakaino, ``Motion generation for food topping challenge 2024: serving
  salmon roe bowl and picking fried chicken,'' {\em Advanced Robotics}, vol.~0,
  no.~0, pp.~1--13, 2025.

\bibitem{directteachingkushida2001human}
D.~Kushida, M.~Nakamura, S.~Goto, and N.~Kyura, ``Human direct teaching of
  industrial articulated robot arms based on force-free control,'' {\em
  Artificial Life and Robotics}, vol.~5, pp.~26--32, 2001.

\bibitem{ablettMultimodalForceMatchedImitation2023}
T.~Ablett, O.~Limoyo, A.~Sigal, A.~Jilani, J.~Kelly, K.~Siddiqi, F.~Hogan, and
  G.~Dudek, ``Multimodal and force-matched imitation learning with a
  see-through visuotactile sensor,'' {\em IEEE Transactions on Robotics}, 2024.

\bibitem{fu2024incontextimitationlearningnexttoken}
L.~Fu, H.~Huang, G.~Datta, L.~Y. Chen, W.~C.-H. Panitch, F.~Liu, H.~Li, and
  K.~Goldberg, ``In-context imitation learning via next-token prediction,''
  {\em arXiv preprint arXiv:2408.15980}, 2024.

\bibitem{hayashi_ieeeaccess_2022}
K.~Hayashi, S.~Sakaino, and T.~Tsuji, ``An independently learnable hierarchical
  model for bilateral control-based imitation learning applications,'' {\em
  IEEE Access}, vol.~10, pp.~32766--32781, 2022.

\bibitem{planningtamizi2023review}
M.~G. Tamizi, M.~Yaghoubi, and H.~Najjaran, ``A review of recent trend in
  motion planning of industrial robots,'' {\em International Journal of
  Intelligent Robotics and Applications}, vol.~7, no.~2, pp.~253--274, 2023.

\bibitem{GR00TN1}
NVIDIA: J.~Bjorck, F.~Castañeda, N.~Cherniadev, X.~Da, R.~Ding, L.~J. Fan,
  Y.~Fang, D.~Fox, F.~Hu, S.~Huang, J.~Jang, Z.~Jiang, J.~Kautz, K.~Kundalia,
  L.~Lao, Z.~Li, Z.~Lin, K.~Lin, G.~Liu, E.~Llontop, L.~Magne, A.~Mandlekar,
  A.~Narayan, S.~Nasiriany, S.~Reed, Y.~L. Tan, G.~Wang, Z.~Wang, J.~Wang,
  Q.~Wang, J.~Xiang, Y.~Xie, Y.~Xu, Z.~Xu, S.~Ye, Z.~Yu, A.~Zhang, H.~Zhang,
  Y.~Zhao, R.~Zheng, and Y.~Zhu, ``Gr00t n1: An open foundation model for
  generalist humanoid robots,'' {\em arXiv:2503.14734}, 2025.

\bibitem{kim24openvla}
M.~Kim, K.~Pertsch, S.~Karamcheti, T.~Xiao, A.~Balakrishna, S.~Nair,
  R.~Rafailov, E.~Foster, G.~Lam, P.~Sanketi, Q.~Vuong, T.~Kollar,
  B.~Burchfiel, R.~Tedrake, D.~Sadigh, S.~Levine, P.~Liang, and C.~Finn,
  ``Openvla: An open-source vision-language-action model,'' {\em arXiv preprint
  arXiv:2406.09246}, 2024.

\bibitem{10295965}
H.~Ichiwara, H.~Ito, K.~Yamamoto, H.~Mori, and T.~Ogata, ``Modality attention
  for prediction-based robot motion generation: Improving interpretability and
  robustness of using multi-modality,'' {\em IEEE Robotics and Automation
  Letters}, vol.~8, no.~12, pp.~8271--8278, 2023.

\bibitem{NIPS2015_a1afc58c}
M.~Watter, J.~Springenberg, J.~Boedecker, and M.~Riedmiller, ``Embed to
  control: A locally linear latent dynamics model for control from raw
  images,'' in {\em Advances in Neural Information Processing Systems},
  vol.~28, Curran Associates, Inc., 2015.

\bibitem{ha2018worldmodels}
D.~Ha and J.~Schmidhuber, ``Recurrent world models facilitate policy
  evolution,'' in {\em Advances in Neural Information Processing Systems 31},
  pp.~2451--2463, Curran Associates, Inc., 2018.

\bibitem{9577560}
M.~Jaques, M.~Burke, and T.~Hospedales, ``Newtonianvae: Proportional control
  and goal identification from pixels via physical latent spaces,'' in {\em
  Proceedings of 2021 IEEE/CVF Conference on Computer Vision and Pattern
  Recognition (CVPR)}, pp.~4452--4461, 2021.

\bibitem{9981610}
R.~Okumura, N.~Nishio, and T.~Taniguchi, ``Tactile-sensitive newtonianvae for
  high-accuracy industrial connector insertion,'' in {\em Proceedings of 2022
  IEEE/RSJ International Conference on Intelligent Robots and Systems (IROS)},
  pp.~4625--4631, 2022.

\bibitem{RTOB}
T.~Murakami, F.~Yu, and K.~Ohnishi, ``Torque sensorless control in
  multidegree-of-freedom manipulator,'' {\em IEEE Transactions on Industrial
  Electronics}, vol.~40, no.~2, pp.~259--265, 1993.

\bibitem{rahmatizadeh_2018}
R.~Rahmatizadeh, P.~Abolghasemi, A.~Behal, and L.~B{\"o}l\"oni, ``From virtual
  demonstration to real-world manipulation using lstm and mdn,'' {\em
  Proceedings of the AAAI Conference on Artificial Intelligence}, vol.~32,
  no.~1, 2018.

\bibitem{KingBa15}
D.~Kingma and J.~Ba, ``Adam: A method for stochastic optimization,'' in {\em
  Proceedings of International Conference on Learning Representations (ICLR)},
  2015.

\bibitem{6523093}
J.~Lee, P.~H. Chang, and R.~S. Jamisola, ``Relative impedance control for
  dual-arm robots performing asymmetric bimanual tasks,'' {\em IEEE
  Transactions on Industrial Electronics}, vol.~61, no.~7, pp.~3786--3796,
  2014.

\bibitem{6224651}
M.~Malhotra, E.~Rombokas, E.~Theodorou, E.~Todorov, and Y.~Matsuoka, ``Reduced
  dimensionality control for the act hand,'' in {\em Proceedings of 2012 IEEE
  International Conference on Robotics and Automation}, pp.~5117--5122, 2012.

\end{thebibliography}

\end{document}